# Evolutionary optimization of spatially-distributed multi-sensors placement for indoor surveillance environments with security levels

Luis M. Moreno-Saavedra[a], Vinícius G. Costa[b], Adrián Garrido-Sáez[a], Silvia Jiménez-Fernández[a], J. Antonio Portilla-Figueras[a], Sancho Salcedo-Sanz[a]

[a]*Department of Signal Processing and Communications, Universidad de Alcalá, Alcalá de Henares, 28805, Madrid, Spain*
[b]*Systems and Computer Engineering Program, Federal University of Rio de Janeiro, Rio de Janeiro, 21941-617, Rio de Janeiro, Brazil*

**Abstract**

The surveillance multi-sensor placement is an important optimization problem that consists of positioning several sensors of different types to maximize the coverage of a determined area while minimizing the cost of the deployment. In this work, we tackle a modified version of the problem, consisting of spatially distributed multi-sensor placement for indoor surveillance. Our approach is focused on security surveillance of sensible indoor spaces, such as military installations, where distinct security levels can be considered. We propose an evolutionary algorithm to solve the problem, in which a novel special encoding (integer encoding with binary conversion) and effective initialization have been defined to improve the performance and convergence of the proposed algorithm. We also consider the probability of detection for each surveillance point, which depends on the distance to the sensor at hand, to better model real-life scenarios. We have tested the proposed evolutionary approach in different instances of the problem, varying both size and difficulty, and obtained excellent results in terms of the cost of sensors' placement and convergence time of the algorithm.

## 1. Introduction

Real-time surveillance of sensible indoor spaces such as militarized areas, governmental buildings, and restricted areas in hospitals or airports, among others, has gained significant importance in recent years [1, 2]. Systems for security surveillance in these areas require effective sensors, sometimes on the framework of the pervasive Internet of Things (IoT) paradigm, plus collaborative decision-making to be fully effective when operating in highly dynamic environments with demanding constraints of time or different security levels [3]. Many existing general surveillance systems are based on a single modality sensor, usually video cameras, sometimes augmented with audio [4]. These systems, however, may not be enough for specific surveillance problems related to security in sensible areas, such as threat detection, specific areas monitoring, or detection of prohibited or unusual events, among others. In these cases, systems based on diverse sensors (video cameras, microphones, mobility sensors, heat detectors, among others) are much more effective and used [5].

In this context, effective sensor placement in indoor surveillance and security problems is critical to obtain robust and effective systems [6]. The optimal location and placement of sensors to form sensor networks have been tackled in different contexts for over 20 years [7, 8, 9]. For instance, the work in [10] was pioneering in defining the Sensor Placement Problem (SPP) with constraints and objectives related to obtaining the minimum possible deployment cost. That work proposed a discrete grid for the available locations and proved different theorems related to the minimum number of sensors necessary to fully cover a given area under the constraints considered. Note that there are also classical problems and results completely related problems to the SPP, such as the alarm placement problem [11] or the guard placement problem in an art gallery [12], which solve linear simpler versions of the SPP. From these initial and classical problems, there have been a large number of studies describing approaches to different versions of the SPP for indoor environments [13], for example considering obstacles in the problem definition [14, 15, 16], mobile sensors [1], or hybrid/multi-sensors [17, 18].

Regarding the main existing computational approaches proposed for tackling SPPs and related problems, heuristics and meta-heuristics algorithms have been some of the most applied methods. Specifically, evolutionary optimization algorithms have been intensively tested in SPPs in surveillance-based applications. In [19] a problem related to automated or semi-automated surveillance monitoring was tackled, consisting of sensor deployment considering coverage needs and overlapping of coverage area. A meta-heuristic solution using a multi-objective evolutionary algorithm was proposed in this case. The paradigm of multi-objective evolutionary computation was also exploited more recently [20] to tackle a problem related to site location optimization to install surveillance cameras. Also, in [21], a problem of sensor placement optimization problem is solved, with application in indoor positioning. The task is solved as a multi-objective optimization problem, where the NSGA-II evolutionary approach was successfully applied. In [22], a genetic algorithm with local search is proposed to solve the problem of sensor deployment optimization related to surveillance applications. Other evolutionary computation approaches for dealing with different problems of optimal sensor placement in wireless networks have been proposed in the last few years [23, 24].

In this paper, we tackle the problem of spatially distributed multi-sensor placement for indoor surveillance, a modern version of the SPP problem. Our approach is focused on security surveillance of sensible indoor spaces, such as military installations or government buildings, where security levels can be associated with points to be monitored, in such a way that a given point must be monitored by several sensors to cope with the corresponding security level. We then consider several different sensor types that must be integrated into the deployment, providing distinct monitoring or surveillance capabilities at diverse costs. The final sensor deployment must fulfill the security level to be a valid solution, and the objective is to obtain a valid solution with the minimum possible cost. We propose an evolutionary-type algorithm to solve the problem with a specific novel encoding (integer encoding with binary conversion) and effective initialization, which speeds up the convergence of the proposed algorithm. We also consider the probability of detection for each surveillance point, which depends on the distance to the sensor. This probability of detection tries to better model real problems, where accurate detection by a sensor is more difficult for long distances. We will show that the proposed evolutionary approach can obtain valid solutions, minimizing the cost of the whole sensor deployment for different instances of the SPP problem considered.

The remainder of the paper has been structured in the following way: the next section presents the problem definition, with details on the problem encoding and specific objective function considered in this work. Section 3 presents the algorithmic approaches developed in this paper, including the EA and recursive algorithm (greedy approach), which serves to initialize the EA. Section 4 describes the experiments carried out in different scenarios of the sensor placement problem considered and the results obtained with the algorithms proposed. Finally, Section 5 presents some conclusions and final remarks from this work.

## 2. Problem definition

In this section, we present the problem definition considered in this paper. It is an SPP with multiple spatially distributed sensors and security levels. The goal is to satisfy all security requirements of a scenario while minimizing the total cost of the whole deployment, i.e., the amount of money spent on all sensor placement, covering all the surveillance requirements. These surveillance requirements are to monitor a set of points, called *Surveillance Points*, that need to be monitored by several types of security sensors, such as high-quality video cameras, smoke sensors or presence, among others. Each surveillance point has different surveillance requirements, e.g., for instance, a given point that has to be monitored by a video camera and a smoke sensor, another point that only has to be monitored by a smoke sensor, and a third point without any need for surveillance, among others. The problem is considered solved when all surveillance requirements are fulfilled. The surveillance security sensors can be dispatched in another set of points called *Sensor Points*. Figure 1 shows a typical scenario used in this problem, where Figure 1a shows the *Surveillance Points* as blue points, Figure 1b shows the *Sensor Points* as red points, Figure 1c shows the scenario with both types of points depicted together and Figure 1d shows the security requirements for each *Surveillance Point*, where each color meaning is the following:

- Sec_0 (green color) points that need to be monitored by 0 sensors (no security requirement for these points is needed).

- Sec_1 (yellow color) points that need to be monitored by one sensor.

- Sec_2 (orange color) points that need to be monitored by two sensors.

- Sec_3 (red color) points that need to be monitored by three or more sensors.

In addition to the characteristics of the points to be monitored, we consider different parameters for each sensor [25], which can be defined with three parameters: the surveillance range of the sensor (measured in meters), its operating angle (measured in degrees), and its cost (measured in Euros).

A solution to the problem consists then of the type and position of all the sensors in the deployment in such a way that they are able to fulfill the surveillance requirements in the scenario, minimizing the total cost of the deployment.

### 2.1. Problem encoding

The encoding considered for this new sensor placement optimization problem consists of an *integer encoding* (each vector value in the encoding is an integer number), with binary conversion to deal with the multi-sensor placement at different points. It is a novel way of encoding the problem and, therefore, a contribution to the work. The details of this encoding are the following: each solution is encoded as an integer vector of size $2N_{sp}$, where $N_{sp}$ stands for the total number of points where one or more sensors can be placed. Thus, the first $N_{sp}$ elements of the vector indicate the surveillance sensors placed at each *Sensor Point*, and the second $N_{sp}$ elements encode the angle coverage of the cameras. Mathematically, this can be expressed in the form $x_{i,j}$, where $i$ indicates the half part of the vector (1 for the first half, 2 for the second half), and $j$ indicates the particular point among the total $N_{sp}$:

$$\mathbf{x} = (x_{1,1}, x_{1,2}, \ldots, x_{1,N_{sp}} \mid x_{2,1}, x_{2,2}, \ldots, x_{2,N_{sp}}), \quad (1)$$

where $x_{1,1}, x_{1,2}, \ldots, x_{1,N_{sp}} \to \left[0, 2^{N_{sen}} - 1\right]$, $N_{sen}$ is the number of surveillance sensors considered, and $x_{2,1}, x_{2,2}, \ldots, x_{2,N_{sp}} \to$

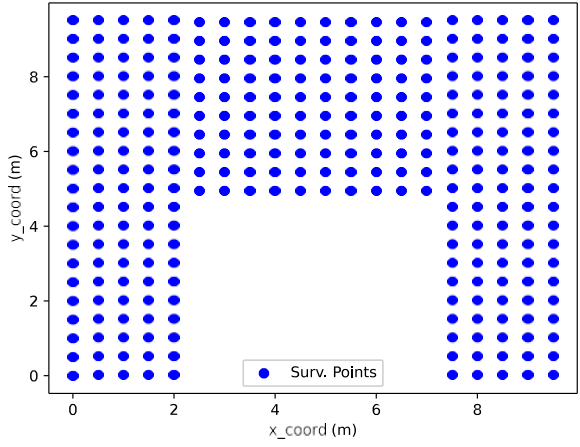
(a) Surveillance Points.

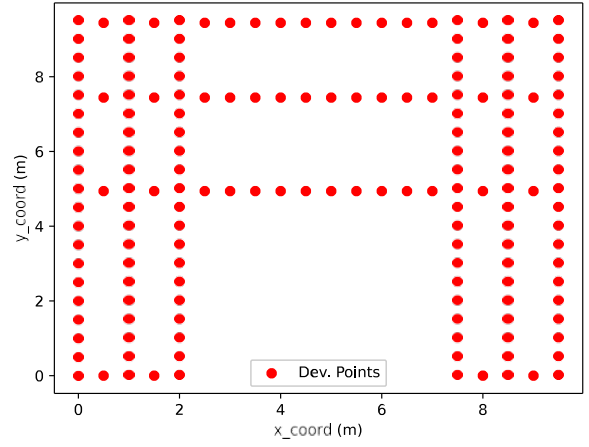
(b) Sensor Points.

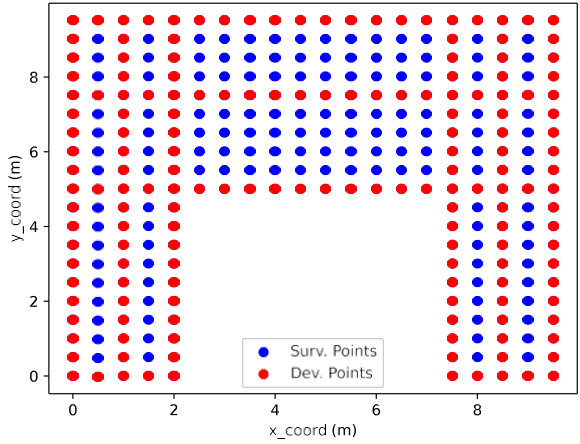
(c) Surveillance and Sensor Points.

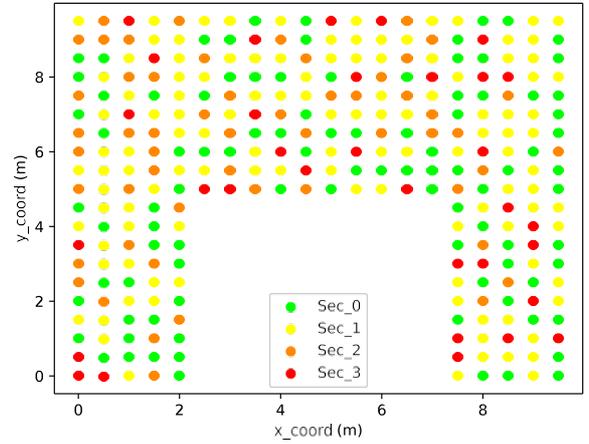
(d) Security Level.

Figure 1: Example of a typical scenario.

[0, 359] stand for the angle coverage, in the case that the sensor is a camera.

In order to finally set what specific sensors are placed at each *Sensor Point* (first part of the encoding), each integer value from the first half of the vector **x** is converted to a binary code, so that the length of each binary number is equal to the number of sensors, and each digit of that binary number indicates whether a sensor is placed at a sensor point (1), or not (0). Figure 2 shows the diagram of integer to binary conversion, where $x$ stands for an integer value in the interval $[0, 2^{N_{sen}} - 1]$ corresponding to a value $x_{1,j}$ (not to be mistaken with the whole vector **x** of Equation (1)).

An example of this encoding is shown below: if we consider a total of 5 surveillance sensors, and the integer number in the encoding is 17, the binary code is 10001, indicating that sensor 1 and sensor 5 are placed at this particular point.

The second half of the vector indicates the orientation of all the sensors placed at each *Sensor Point*, also using an integer encoding. Each integer value of the second half of vector **x** indicates, in this case, the orientation of the device, in decimal degrees, where $0^{\downarrow}$ is up, $90^{\downarrow}$ is right, $180^{\downarrow}$ is down, and $270^{\downarrow}$ is left.

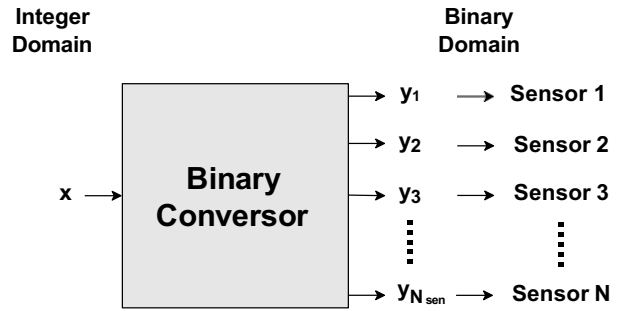

Figure 2: Integer to binary conversion representation. $x$ in the figure stands for an integer value, and $y_1$, $y_2$, $y_3$, $y_4$ and $y_{N_{sen}}$ the digits of the binary conversion.

Figure 3 shows an example of the encoding in a supposed case with 4 Sensor points. In this example, it can be seen that number 17 in integer encoding is converted to the binary number [1 0 0 0 1], which means that sensors 1 and 5 are placed at this point (point 1) with an orientation of 257°. The number 12 in the encoding is converted to the binary number [0 1 1 0 0], which means sensors 2 and 3 are placed at point 2, with an orientation of 0°. In turn, the number 0 in the encoding is converted to the binary number [0 0 0 0 0], which means that no sensor is placed at point 3, and, finally, the number 5 in the encoding is converted to the binary [0 0 1 0 1], which means that sensors 3 and 5 are placed at point 4, with an orientation of 135°. Note that this integer encoding with binary transformation allows us to deal with multiple sensors and also consider the angle coverage of the sensor devices at each point.

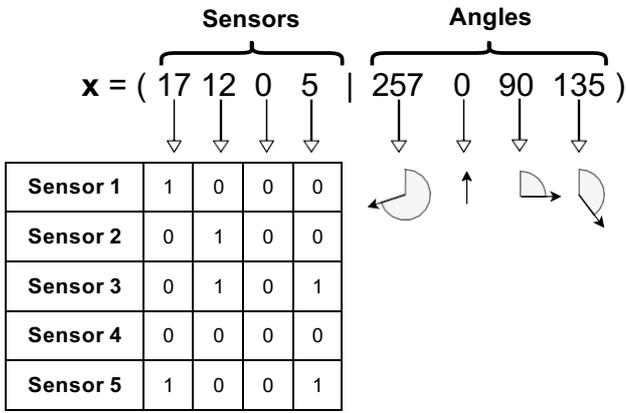

Figure 3: Integer encoding example. The first part of the vector indicates the sensor deployment, and the second part indicates the orientation angle of all devices at each point.

*2.2. Objective function*

We define the objective function of this problem as the cost of the whole deployment (placing of the necessary sensors that solve the problem) in Euros. Equation (2) shows the total cost of the deployment:

$$g(\mathbf{x}) = \sum_{i=1}^{N_{sen}} C_i \cdot N_i + P, \quad (2)$$

where $g(\mathbf{x})$ is the fitness of each individual, measured in euros, $P$ is a penalty term if the surveillance requirements are not fulfilled, $N_{sen}$ is the total types of sensors, $C_i$ is the cost of each sensor and $N_i$ is the number of that particular sensor. The penalty term consists of 1.000.000 for each *Surveillance Point* that is not monitored by the required sensors, so penalized solutions will be discarded by the algorithm in the following evolution steps.

To calculate the objective or fitness function, we need to calculate firstly three different matrices: distance matrix, angle matrix, and vision matrix, all with $N_{sp} \leftrightarrow N_{surv}$ dimension, where $N_{sp}$ is the number of *Sensor Points* and $N_{surv}$ the number of *Surveillance Points*. Distance matrix, $M_D$, is a matrix where each element $d_{ij}$ stands for the distance, in meters, between the *Sensor Point i* and the *Surveillance Point j*. Angle matrix, $M_A$, is a matrix where each element $a_{ij}$ stands for the angle (decimal) between *Sensor Point i* and *Surveillance Point j*, and finally, Vision matrix, $M_V$, is a matrix where each element $v_{ij}$ is the direct vision between *Sensor Point i* and *Surveillance Point j*, with the value 1 if direct vision exists between them and value 0 otherwise.

For each *Sensor Point* in the solution individual, the integer number is decoded as shown in Figure 2. In those sensors with value 1, i.e., in the case that the sensor is placed at the *Sensor Point*, we extract its maximum distance and angle values, and we compare those values with the specific row in Distance Matrix $M_D$, in Angle Matrix $M_A$ and Vision Matrix $M_V$.

- In the angle case, the values that fail to meet the angle requirements are set to 0, which means that the *Surveillance Point* is not covered by that sensor, and they are set to 1 otherwise, obtaining the covered *Surveillance Points* by angle. With all the sensors and all the *Surveillance Points*, we build a new matrix, the points covered by the angle with $N_{surv} \leftrightarrow N_{sen}$ dimension and called $M_{CPA}$.

- To better model real cases where the distance from the point to be monitored and the sensor plays a central role in the detection, we introduce a probability of detection that depends on the distance. This way, the distance between the sensor and the *Surveillance Point* is transformed into a detection probability, $P_D$, using the sigmoid function, as expressed in Equation (3) [26]:

$$P_D(d) = \frac{-1}{1 + e^{-0.5(d-Max_{dist})}} + 1 \quad (3)$$

Using Equation (3), we obtain a $P_D$ for each *Surveillance Point* and each sensor deployed, where $Max_{dist}$ is the maximum distance range for each sensor, measured in meters. Figure 4 shows the detection probability considering a linear case (blue line) with crisp detection at $Max_{dist}$, and the non-linear case (red line), used in this work, which allows a softer detection centered at $Max_{dist}$. To calculate the total probability of detection [27] for each sensor in each *Surveillance Point*, we use Equation (4):

$$P_{D_T} = 1 - \prod_{i=1}^{N_{sen}} (1 - P_{D_i}), \quad (4)$$

where $P_{D_T}$ is the total probability of detection, $N_{sen}$ is the number of sensors and $P_{D_i}$ is the probability of detection calculated using Equation (3). Finally, we build the matrix $M_{CPD}$ by establishing a threshold over the detection probability calculated above, with a value of 0 if the *Surveillance Point* is not covered for distance by the sensor (its value is lower than the threshold set), and a value of 1 otherwise.

- In the direct vision case, we build a new matrix, $M_{CPV}$, with the same dimension as $M_{CPA}$ and $M_{CPD}$, and containing a 1 if a *Surveillance Point* is covered with each sensor type, and a 0 otherwise.

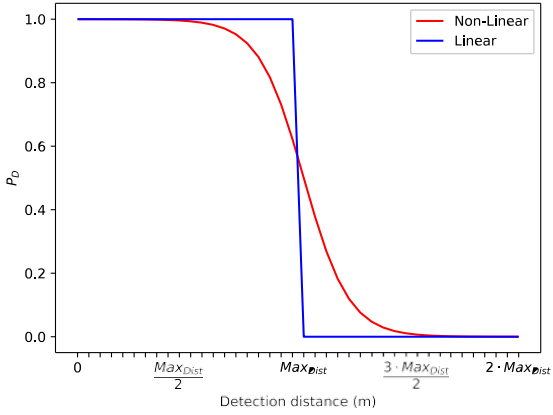

Figure 4: Detection probability of a given sensor with maximum coverage $Max_{dist}$. Crisp detection (blue line) and Soft detection (red line) used in this work.

The final monitored points are obtained then by multiplying the matrices calculated above, i.e., $M_{CPA}$, $M_{CPD}$ and $M_{CPV}$, obtaining the Covered Points in Total Matrix ($M_{CPT}$), as shown in Equation (5):

$$M_{CPT} = M_{CPA} \cdot M_{CPD} \cdot M_{CPV} \qquad (5)$$

The total number of points not covered in the deployment, which will lead to a penalty term in the algorithm, is calculated by obtaining the *Surveillance Points* that have to be monitored and the total covered points (with Equation (5)). Table 1 shows a truth table to get the total points not covered in the deployment, and then a penalization term must be applied to the individuals in the algorithm.

Table 1: Output truth table (O), where a 1 value indicates there are no errors, 0 otherwise. A indicates if a Surveillance Point needs to be monitored (1) or not (0). B indicates if a Surveillance Point is already monitored (1) or not (0).

| A | B | O |
|---|---|---|
| 0 | 0 | 1 |
| 0 | 1 | 1 |
| 1 | 0 | 0 |
| 1 | 1 | 1 |

In Table 1, the first column (A) stands for the need to be monitored for each *Surveillance Point*, where 0 indicates that point does not need to be monitored, and 1 otherwise. The second column (B) indicates if the point is monitored (with 1 value) or if it is not monitored (with value 0). The third column (O) is the output, where 1 value indicates there is no failure, and 0 indicates a failure, so the penalization has to be applied. The only case where the penalization is applied is where a *Surveillance Points* must be monitored and it is not monitored, the other cases are correct:

- The first row represents the case when a *Surveillance Points* must not be monitored, and it is not monitored.

- The second row indicates that a *Surveillance Points* must not be monitored, but it is monitored.

- The last row occurs when a *Surveillance Points* must be monitored, and it is indeed correctly monitored.

Equation (6) shows the logic operation to get the output:

$$Output = \overline{A} + B \qquad (6)$$

Figure 5 shows an example of the same scenario shown in Figure 1, but indicating the monitored points by distance restriction (Figure 5a), angle restriction (Figure 5b), vision restriction (Figure 5c) and total monitored points (Figure 5d). The red point in the figures is the point where the sensor is deployed, the green points are points that are monitored by the sensor, and the blue points are points that are not covered by the sensor.

## 3. Proposed algorithmic approaches

### 3.1. Recursive Algorithm

In this section, we present a Recursive Algorithm (RA) to solve the problem. It is a constructive heuristic whose objective is to obtain a good enough solution, which can then be mutated and used in the Evolutionary Algorithm population initialization to reduce the algorithm's convergence time and improve the performance.

The recursive algorithm proposed starts by considering all the *Surveillance Points* that have to be monitored, but without any surveillance (starting of the algorithm) for each sensor. Randomly, an unmonitored point is chosen, and a sensor is placed near this point to ensure that the point is now monitored. Note that this sensor does not cover this point exclusively, so other points are also monitored by including this sensor. Next, another unmonitored point is chosen, and the same procedure is carried out. When all the points are covered for each sensor, the algorithm is stopped. At the start of the algorithm, the angle where each sensor is deployed is also randomly generated. Algorithm 1 shows the complete algorithm explained:

---
**Algorithm 1:** Recursive Sensor Placement Algorithm

**Input:** Set of Surveillance Points $N_{sp}$, number of sensors $N_{sen}$

**Output:** $x$

1 Randomly generate initial angles for each sensor;
2 Actual sensor $S_i \leftarrow 0$;
3 $x \leftarrow \emptyset$;
4 **while** $\exists$ *unmonitored point in $N_{sp}$* **do**
5     Randomly select an unmonitored point $p$ from $N_{sp}$;
6     Place a sensor near $p$ ensuring coverage of nearby points;
7     Update $x$ with the placed sensor;
8 **end**
9 **return** $x$;
---

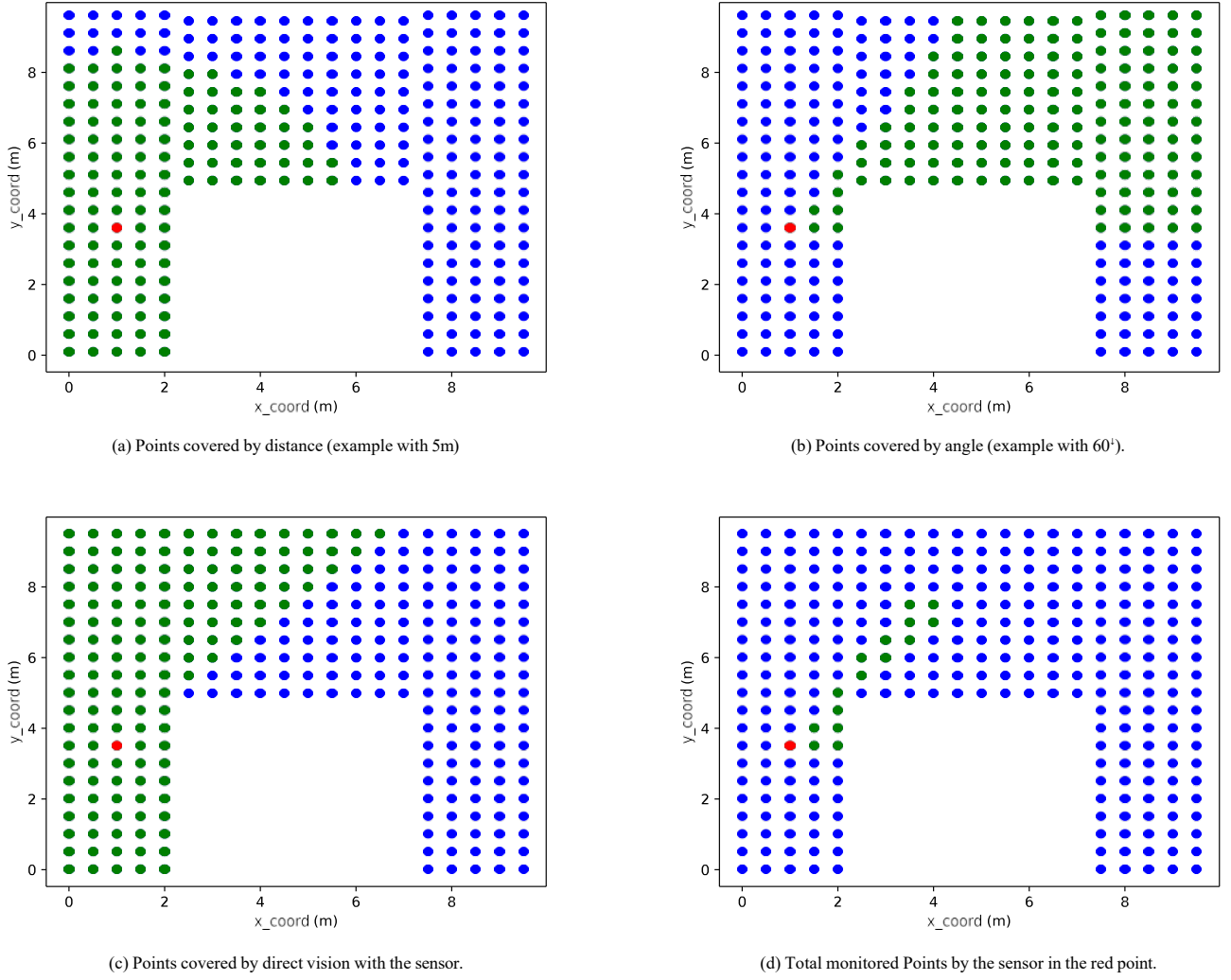

Figure 5: Monitored points by a sensor set in the position of red point, green points are the monitored points, and blue points represent the uncovered points.

### 3.2. Greedy randomized adaptive search procedure (GRASP)

The Greedy Randomized Adaptive Search Procedure Algorithm (GRASP) [28, 29] is a meta-heuristic that combines elements of local search with a greedy approach to solve optimization problems.

The main idea behind GRASP is to generate multiple greedy solutions to the problem and apply several iterations of a local search for each greedy solution. In this work, the greedy solutions are generated using the RA explained in Section 3.1, and the local search is carried out using different mutation operators, as follows:

- Randomly select a gene and increase its value by 1 unit (in sensors vector) and by 30 units (in angles vector).

- Randomly select a gene and decrease its value by 1 unit (in sensors vector) and by 30 units (in angles vector).

- Randomly select a gene and replace the value with an integer random number in the interval $[0, 2^{N_{sen}} - 1]$ in sensors vector and other integer value in the interval $[0, 359]$ in angles vector.

- Randomly select a gene and set its value to 0.

Algorithm 2 shows the complete algorithm explained:

### 3.3. Proposed Evolutionary Algorithm

In this section, we describe the proposed Evolutionary Algorithm with Integer Encoding (EA-IE) [30, 31] to solve the SPP proposed in this work. Evolutionary Algorithms (EAs) are a kind of optimization method inspired by Darwin's theory of evolution. They work on the idea that the strongest and most adaptable individuals in a population are more likely to survive and pass on their traits to the next generations. This process of natural selection helps improve solutions over successive iterations, favoring the perpetuation of the most effective characteristics for a particular environment. The principle of the EA is to mimic the behavior of living things, using mechanisms

**Algorithm 2:** Greedy Randomized Adaptive Search Procedure (GRASP)

**Input:** Number of Greedy algorithm iterations $I_{GR}$, number of Local Search iterations $I_{LS}$
**Output:** Best solution $x_{best}$, Population of best solutions $P_{best}$

1 $iter_{GR} \nearrow 0$;
2 $iter_{LS} \nearrow 0$;
3 **while** $iter_{GR} < I_{GR}$ **do**
4     $x_{best,i} \nearrow$ GenerateGreedySolution();
5     **while** $iter_{LS} < I_{LS}$ **do**
6        $x \nearrow$ LocalSearch($x_{best,i}$);
7        **if** $x$ is better than $x_{best,i}$ **then**
8           $x_{best,i} \nearrow x$;
9        **end**
10        $iter_{LS} \nearrow iter_{LS} + 1$;
11    **end**
12    $iter_{GR} \nearrow iter_{GR} + 1$;
13    $P_{best} \nearrow x_{best,i}$;
14 **end**
15 $x_{best} \nearrow \min(P_{best})$;
16 **return** $x_{best}$;

associated with biological evolution, e.g., reproduction, mutation, or recombination, which are implemented in a computer to solve optimization problems. Thus, from an algorithmic point of view, an EA is an iterative optimization algorithm that uses this principle to explore global solutions. Figure 6 shows the workflow of the proposed EA with integer encoding.

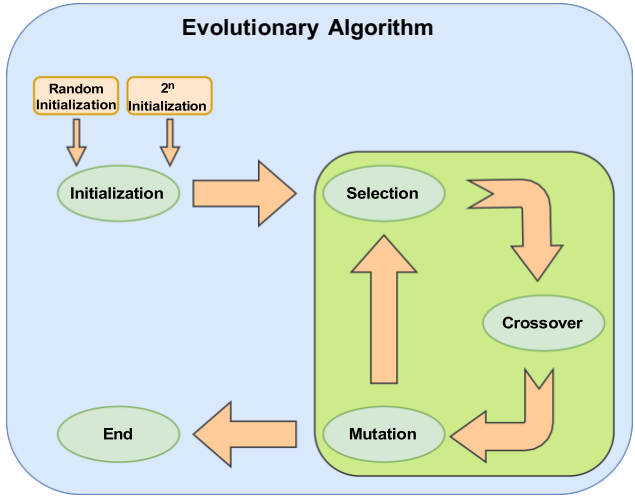

Figure 6: Workflow of the Evolutionary Algorithm for the considered SPP.

As shown in Figure 6, the proposed EA algorithm starts from a set of initial solutions, encoded according to the integer encoding and binary conversion described above and sorted according to their associated costs using the fitness function of the problem, also explained in the previous section. In this case, half of the solutions are randomly initialized, and the rest are initialized using a special initialization, where values of the form $2^n$ are more taken into account. This initialization is due to the particular integer encoding system where $2^n$ values mean only one sensor is deployed at this specific point. The top-performing solutions within the population, i.e., those with the lowest cost when minimizing the fitness function and those with the highest cost when maximizing the fitness function, have a higher chance of preserving their features in the next generation. That occurs because they are more likely to be chosen for reproduction through a process called the *Crossover Operation*, which generates new solutions. In the crossover operation, the features of two parents are mixed up to produce new individuals for the population, forming a new offspring. In addition, each time these new individuals are generated, there is the possibility that they may undergo random alterations or mutations in the EA, in a process known as *Mutation Operation*. It consists of taking one or more features of the individual and modifying the value by another random one, allowing the algorithm to escape from local minima. New fitness value costs are assigned to the offspring individuals and compared with the original ones in the population. The individuals with the worst costs in the population are discarded in a process called *Selection operator*. This process is repeated until the maximum number of iterations is reached or until convergence. Algorithm 3 also describes the EA workflow in detail.

**Algorithm 3:** Workflow of the EA algorithm.

**Input:** Population $P$, population size $N$, number of generations $G$
**Output:** Best solution $x_{best}$

1 $t \nearrow 0$;
2 $P_0 \nearrow$ Initial population generation;
3 **while** $t < G$ **do**
4     $P_t \nearrow$ Selection operations between individuals from $P_{t-1}$;
5     $P_t \nearrow$ Crossover operation between individuals in $P_t$;
6     $P_t \nearrow$ Mutation operation of some new individuals after Crossover operation in $P_t$;
7     $x_{best} \nearrow$ Best individual in $P_t$;
8     $t \nearrow t + 1$;
9 **end**
10 **return** $x_{best}$;

The operators implemented in the Evolutionary Algorithm are the following:

- Initialization Operation: At the start of the Evolutionary Algorithm, the population, i.e., the set of solutions, is formed. In this case, we develop two different ways for the initialization:
  - Random Initialization: Each individual of the population is formed using a random uniform distribution
  - $2^n$ Initialization: Each problem solution is formed by multiples of 2, e.g., 1, 2, 4, 8, among others.

That is due to the problem encoding explained in Section 2.1. The multiples of two in binary means that only one sensor is deployed in the *Surveillance Point*.

- Selection Operation: The selection operation is performed when a generation of EA starts. In the selection operation, only the individuals best adapted to the environment survive, i.e., the solutions with lower objective function values in the minimization case and solutions with higher objective function values in the maximization case.

- Crossover Operation: Once the selection operation has been carried out, the next operation is the crossover operation. We implement in this problem a multi-point crossover that consists of generating a new individual ($\mathbf{y}$) by randomly selecting some genes from one individual ($\mathbf{x_1}$) and the rest of the genes from the other individual ($\mathbf{x_2}$). Figure 7 shows an example of a multi-point crossover operation.

| $\mathbf{x_1}$ | 2 | 3 | 5 | 1 | 0 | 2 | 1 | 4 |
| $\mathbf{x_2}$ | 4 | 2 | 3 | 2 | 4 | 1 | 3 | 0 |
| $\mathbf{y}$ | 2 | 2 | 5 | 1 | 4 | 1 | 1 | 4 |

Figure 7: EA-IE crossover operation. $\mathbf{y}$ is the resulting solution after the crossover.

- Mutation Operation: Each time a new individual is generated in the crossover operation, there is a possibility that this individual will mutate. We use two different types of mutation in this Evolutionary Algorithm. The first mutation operation consists of replacing one of the gene values with a random value. Figure 8 shows an example of the mutation operation implemented in the EA. The second mutation operation is like the other mutation, but the value replaced is set to 0. That helps the algorithm eliminate unnecessary sensors placed in other steps.

| $\mathbf{x}$ | 2 | 3 | 5 | 1 | 0 | 2 | 1 | 4 |
| $\mathbf{y}$ | 2 | 3 | 5 | 1 | 0 | 1 | 1 | 4 |

Figure 8: EA-IE mutation operator. $\mathbf{y}$ is the resulting solution after the mutation.

*3.4. Initialization with the recursive algorithm*

The final initialization of the proposed EA is obtained by considering solutions from the RA described above (Section 3.1). The RA is first used to get some initial solutions (not extremely good) to help the EA converge faster. The initialization in the EA when the recursive algorithm is considered is as follows:

- 33% of the population is randomly set with integer values, as explained in Section 3.3.
- 33% of the population is set using the $2^n$ initialization, as explained in Section 3.3.
- 34% of the population is set using the Recursive Algorithm.

With this special initialization, we get good solutions at the first stages of the EA, better than those obtained with a random algorithm initialization. We also get a good diversity in the initial solutions of the EA, which allows a better exploration of the search space. Finally, note that the inclusion of solutions from the recursive algorithm makes the convergence time and the performance of the EA much better, as we will show in the experimental part of this work.

**4. Experiments and Results**

In this section, we present the data generated and used for each experiment (Section 4.1) and the results of different experiments carried out to show the performance of the algorithms proposed (Section 4.2).

*4.1. Data generation and experiments description*

The data available for the experiments consists of a set of artificially generated scenarios that simulate real-world rooms or buildings to get near real-world cases. All the generated scenarios have two different types of points, i.e., *Surveillance Points* and *Sensor Points* as explained above. Each *Surveillance Point* has a different security level of surveillance, randomly generated, and the *Sensor Points* are generated according to each scenario. There are three different types of scenarios, each one with a different shape and point composition. However, we can group the scenarios into three types by their size, shown in Figure 9:

- Small-size scenarios: These scenarios simulate a room that has to be surveyed in a building. They are the simplest scenarios, composed of a squared grid with size 10 ↔ 10 meters and 20 ↔ 20 points, with zones without points, simulating different shapes. Scenario 1 (Figure 9a), Scenario 2 (Figure 9b), and Scenario 3 (Figure 9c) are the scenarios of this type.

- Medium-size scenarios: These scenarios simulate a set of different rooms in a building. They are formed by merging five small-size scenarios, so they are composed of a squared grid with size 30 ↔ 30 meters and 60 ↔ 60 points, including some zones without points, simulating the floor plan of a building. Scenario 4 (Figure 9d), Scenario 5 (Figure 9e), and Scenario 6 (Figure 9f) are the scenarios of this type.

- Large-size scenario: A final large-size scenario is considered, simulating the floor plan of a large industrial building. This scenario is formed using nine small-size scenarios, and it is composed of a squared grid with size 30 ↔ 30 meters and 60 ↔ 60 points. Scenario 7 (Figure 9g) is the only scenario of this type.

The security levels considered for each point in all the scenarios are shown in Figure 10. As can be seen, in all the scenarios contemplated, the number of points with the maximum level of security (red points in the figure), which must be covered by different sensors, is high, which makes the problems associated with these scenarios quite complex.

The surveillance sensors used in each experiment and their specifications are the following: two video cameras with different specifications (one for general surveillance, considering a wide-angle lens, and another one dedicated to monitoring specific areas, with a higher resolution), one volumetric-motion sensor to detect movements, a seismic detector to detect possible breaks through windows or doors, and finally a smoke sensor, to prevent fires. These sensors' specifications are detailed in Table 2.

Table 2: Surveillance sensors used in the experiments and their specifications.

| Sensor | Range(m) | Angle(º) | Cost(e) |
|---|---|---|---|
| Wide-angle lens camera | 30 | 180 | 42 |
| Narrow-angle lens camera | 60 | 30 | 112 |
| Volumetric-motion sensor | 18 | 90 | 42 |
| Seismic detector | 5 | 360 | 169 |
| Smoke detector | 4 | 360 | 6 |

Table 3 shows general information and hyper-parameters values for each algorithm employed in this work to obtain the results shown in Section 4.2. All the algorithms in all the scenarios have been simulated using the same number of fitness function calls (100k) to get a fair comparison between them, and the hyper-parameters have been set according to this restriction.

Table 3: General information and hyper-parameter values in the algorithms

| General information | |
|---|---|
| Number of simulations | 30 |
| Seeds | [2023, 2053] |
| **RA** | |
| Number of greedy iterations | 100000 |
| **GRASP_1** | |
| Number of greedy iterations | 100 |
| Number of local search iterations | 1000 |
| **GRASP_2** | |
| Number of greedy iterations | 500 |
| Number of local search iterations | 200 |
| **EA and EA+RA** | |
| Number of individuals | 100 |
| Number of generations | 1000 |
| Mutation prob. | 0.7 |
| Crossover frac. | 0.5 |
| Survival frac. | 0.5 |

We carried out 30 simulations for each scenario for each algorithm to get significant statistical results, and we used the same fixed seeds for experiment reproducibility. The algorithms employed are the GRASP algorithm (Section 3.2), the EA (Section 3.3), and the mixed approach between RA and EA (Section 3.4). In the case of the GRASP algorithm, we define two different experiments, changing the hyper-parameters to analyze the impact of the number of iterations in greedy and local search algorithms.

*4.2. Results*

Table 4 shows the statistical results obtained, i.e., best simulation (Min.), worst simulation (Max.), average of simulations (Mean), and standard deviation (Std.), in the fitness function value of the 30 simulations for each algorithm in each scenario.

It can be seen that the best algorithm in general terms is the EA+RA, i.e., the hybrid model that initializes a part of the population using the RA and then lets the EA start its workflow. Regarding the comparison of the two variants of the GRASP algorithm, it can be seen that in all cases, GRASP 1 consistently outperforms GRASP 2, indicating the best choice of hyperparameters is to assign more importance to local search than to the creation of greedy solutions for this particular problem.

A comparison of the three algorithms according to the type of scenarios reveals that in the small-size scenarios, i.e., scenarios 1, 2, and 3, the results are very similar. This is evidenced by minimal differences between the algorithms. However, it is interesting that in none of the scenarios, the final approach (EA+RA) gets the best results, although its results are good. This lack of differentiation may be attributed to the relatively simple nature of these scenarios, which may not be sufficiently challenging to distinguish between the algorithms. In the medium-sized scenarios, i.e., scenarios 4, 5, and 6, the GRASP algorithm begins to exhibit inferior performance relative to the EAs, thereby making the GRASP algorithm an unsuitable choice for tackling medium-sized scenarios of this nature. In addition, the EA+RA algorithm is already able to achieve superior results compared to its competitors. In the case of the largest scenario, the EA+RA algorithm achieves significantly better results than the rest of the approaches compared. Note, however, that a comparison of the GRASP algorithm with the EA algorithm (randomly initialized) in this scenario reveals that the GRASP algorithm achieves superior results, so in this case, the EA initialization with the RA plays an important role in improving the EA convergence, which is not found when the algorithm is randomly initialized.

We also show here a comparison of the different EAs considered regarding aspects such as convergence time or the evolution of each of the solutions throughout the generations in all the scenarios. Figure 11 shows the fitness function cost evolution along the generations in the different scenarios considered. In the blue color line, we depict the EA evolution and, in orange, the combination of EA and the RA initialization. It can be seen that the combination of the EA and the RA can obtain better results using less computation time. In medium-size scenarios, the best fitness function cost in the final ensemble is obtained approximately in generation 600, while in the EA without initialization, this fitness is obtained at the end of the algorithm. In the most complex scenario, i.e., Figure 11g, the performance of the final ensemble is pretty superior, and the EA is not able to get an optimal result in all the experiment duration, while the final ensemble approach has already converged to an optimized solution of the problem.

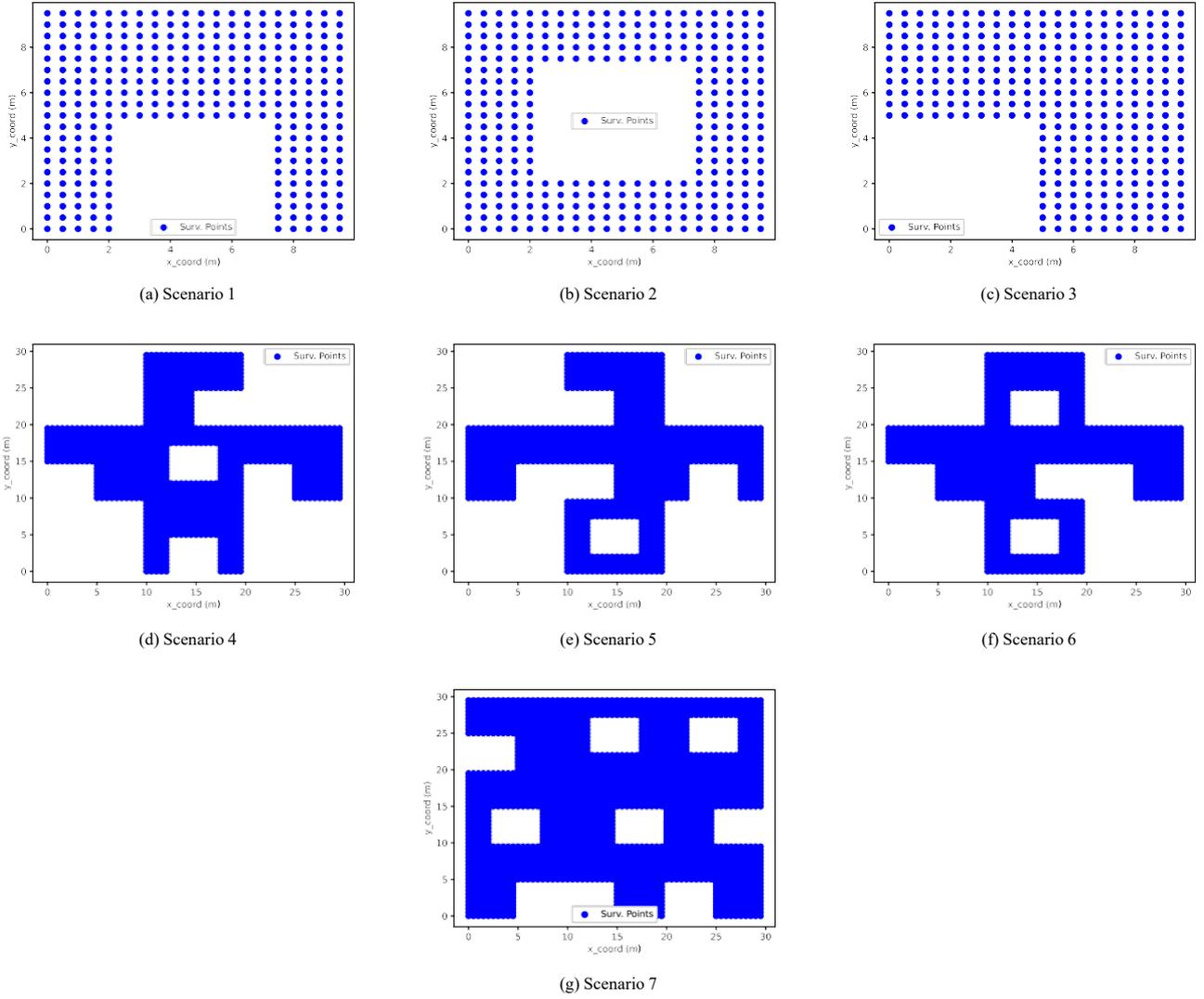

Figure 9: Surveillance points for each scenario considered, categorized as small-size scenarios (9a, 9b and 9c), medium-size scenarios (9d, 9e and 9f) and large-size scenario (9g).

Finally, Figure 12 shows the representation of the best solutions obtained in three scenarios (small, medium, and big size), detailed for each sensor, where it can be seen the monitored area in yellow color. Blue points are those that do not need to be watched for that sensor, and red points are those that have to be monitored. It also can be seen that each sensor has different coverage ranges and angles, and there are no red points without monitoring in any case, fulfilling the security constraints of the problem.

## 5. Conclusions

In this paper, we have dealt with a spatially distributed multi-sensor placement problem for indoor surveillance applications, considering different security levels. The objective of the problem is to obtain a feasible deployment of sensors fulfilling the security requirements, which minimizes the deployment cost. A probability of detection for each surveillance point, which depends on the distance to the sensor, has also been considered to better simulate real cases of application. We have proposed an integer encoding with a binary conversion Evolutionary algorithm, which can manage the multi-sensor characteristic or the considered sensor placement problem. A recursive algorithm to generate good-enough solutions has been implemented and hybridized with the EA to obtain better initializations of the population, with pretty good effects over the EA performance, mainly in large-size problems. We have also tested the performance of a Greedy Randomized Adaptive Search Procedure (GRASP) to this problem.

The performance of the proposed EA with random and recursive initializations has been tested in different constructed scenarios of diverse sizes (small, medium, and large), which simulate rooms and building plants with several security requirements. Results have shown that the EA can solve the problem, obtaining good-quality solutions in all cases studied. The experiments carried out have shown that the recursive initialization is positive in medium and large scenarios, where the random initialization produces a slow convergence of the EA, which can be fixed by using the recursive initialization. However, the results in small-size scenarios have shown that this initialization leads the EA to sub-optimal search space parts, so after a fast convergence in the first stages of the algorithm, the EA cannot move to alternative zones of the search space with better solutions. The EA using random solutions starts with much worse individuals, but then it can improve the search and cover larger parts of the search space to finally obtain better results in the small-size scenarios. This effect has not been detected in medium or large-size scenarios, where the initial improvement of the recursive initialization is enough to guarantee a better final solution than the solution obtained using random initialization after the determined number of generations. In a comparison with the GRASP algorithm, we have shown that the GRASP can get better results than the EA with random initialization in the largest scenario, but the EA with recursive initialization can obtain better results, improving the performance of the GRASP.

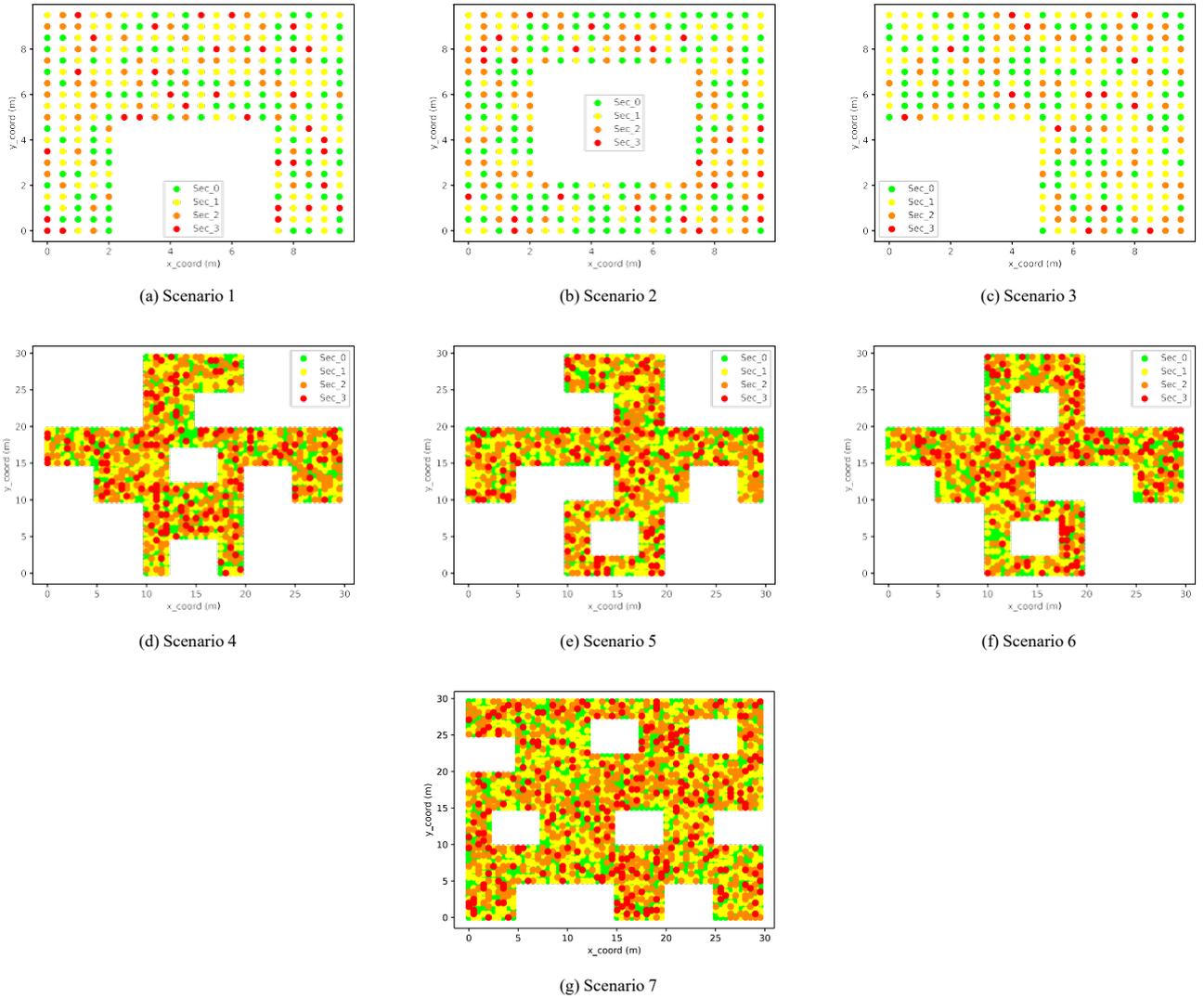

Figure 10: Security levels of all points for each scenario considered.

Future research lines will include the development of novel specific crossover and mutation operators adapted to the problem at hand. These types of specific or tailor-made operators are usually more costly concerning computation but usually improve the algorithm performance if they are well-designed. We also plan to extend these specific crossover and mutation operators to enhance the performance of the GRASP algorithm developed.

Table 4: Results of experiment 2 in all scenarios.

|  |  | GRASP 1 | GRASP 2 | EA | EA+RA |
|---|---|---|---|---|---|
| Scenario 1 | Min. (e) | 1617.0 | 2251.0 | **1427.0** | 1434.0 |
|  | Max. (e) | **1974.0** | 2820.0 | 2189.0 | 2337.0 |
|  | Mean (e) | 1868.23 | 2534.3 | **1722.8** | 1824.4 |
|  | Std. (e) | **83.646** | 136.25 | 195.96 | 233.89 |
| Scenario 2 | Min. (e) | 2035.0 | 2868.0 | **1545.0** | 1651.0 |
|  | Max. (e) | 2590.0 | 3778.0 | **2487.0** | 2730.0 |
|  | Mean (e) | 2354.6 | 3350.63 | **1941.7** | 2143.6 |
|  | Std. (e) | **127.63** | 197.27 | 229.27 | 256.37 |
| Scenario 3 | Min. (e) | 1238.0 | 1774.0 | **1126.0** | 1196.0 |
|  | Max. (e) | **1631.0** | 2116.0 | 1673.0 | 1967.0 |
|  | Mean (e) | 1448.2 | 1964.37 | **1366.0** | 1517.37 |
|  | Std. (e) | **79.254** | 96.64 | 147.42 | 188.34 |
| Scenario 4 | Min. (e) | 7670.0 | 9203.0 | 4932.0 | **4652.0** |
|  | Max. (e) | 9091.0 | 10548.0 | 9764.0 | **5962.0** |
|  | Mean (e) | 8351.5 | 10086.8 | 5933.2 | **5228.9** |
|  | Std. (e) | 338.23 | **287.57** | 983.61 | 352.29 |
| Scenario 5 | Min. (e) | 8758.0 | 10687.0 | 5017.0 | **4791.0** |
|  | Max. (e) | 9921.0 | 11875.0 | **7380.0** | 7598.0 |
|  | Mean (e) | 9437.2 | 11388.2 | 5822.3 | **5779.9** |
|  | Std. (e) | 321.65 | **287.041** | 538.89 | 551.20 |
| Scenario 6 | Min. (e) | 8796.0 | 10413.0 | 5003.0 | **4906.0** |
|  | Max. (e) | 10049.0 | 11905.0 | 7151.0 | **6563.0** |
|  | Mean (e) | 9472.9 | 11288.5 | 5889.6 | **5636.6** |
|  | Std. (e) | **317.05** | 427.31 | 503.52 | 383.17 |
| Scenario 7 | Min. (e) | 12258.0 | 14071.0 | 14768.0 | **6191.0** |
|  | Max. (e) | 14063.0 | 15711.0 | 21340.0 | **8724.0** |
|  | Mean (e) | 13280.5 | 14986.8 | 17820.6 | **7092.7** |
|  | Std. (e) | 463.89 | **395.82** | 1658.28 | 542.95 |

**Acknowledgements**

This work is supported by the project PID2020-115454GB-C21 of the Spanish Ministry of Science and Innovation (MICINN). This research has also been partially supported by Universidad de Alcalá - ISDEFE Research Chair in ICT and Artificial Intelligence, and by the Brazilian research agency Coordination of Higher Education Personnel Improvement - Finance Code 001. We would like to especially thank Major Román Barroso from the Joint Cyberspace Command (MCCE) for their support and contribution to this research.

**Data and software**

All the data (scenarios) and software developed (EA+RA) are available at:
https://github.com/LuisMiguelMoreno/multisensor-deployment

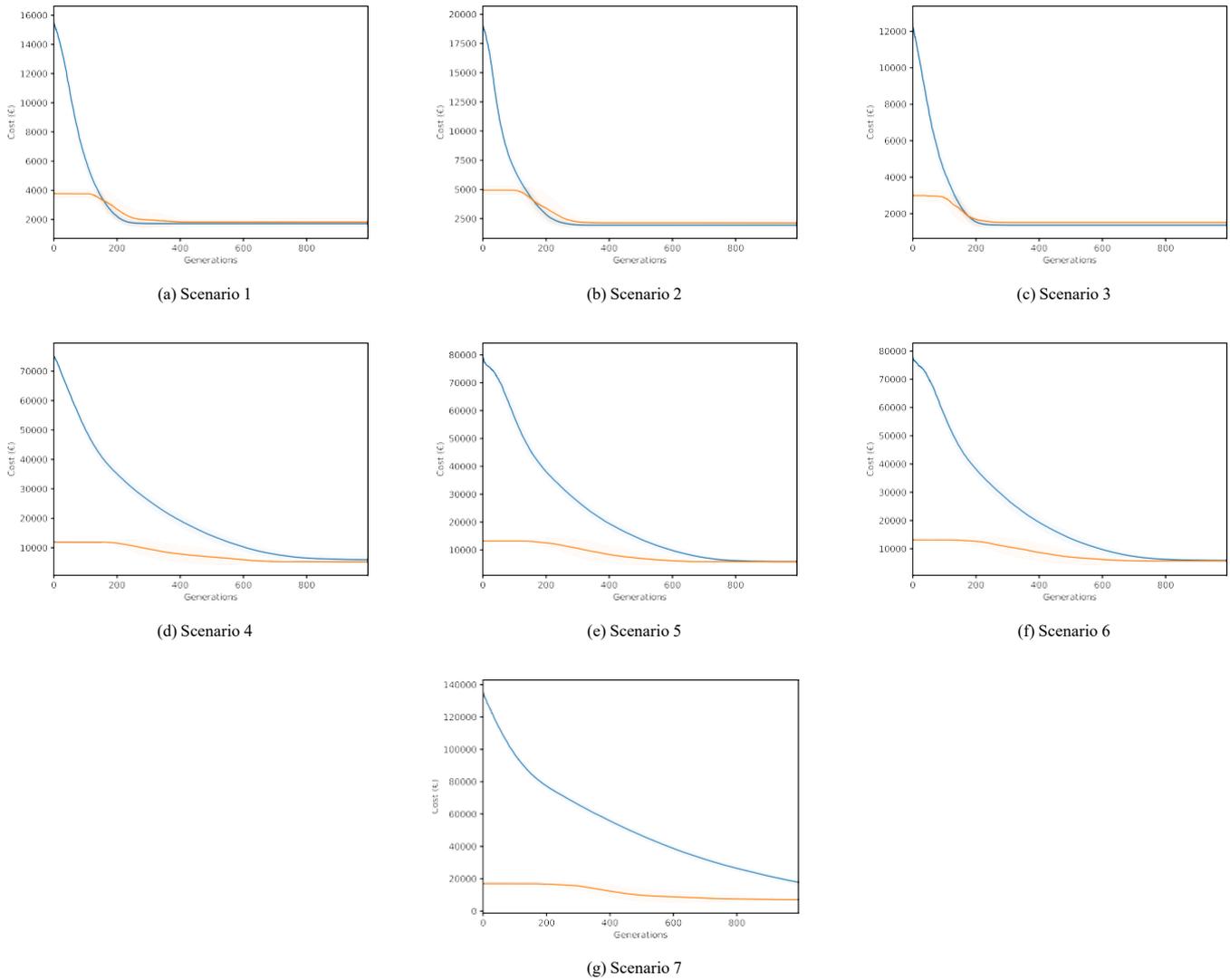

Figure 11: Fitness function comparison between Evolutionary Algorithm (blue line) and Evolutionary Algorithm with Recursive Algorithm Initialization (orange line)

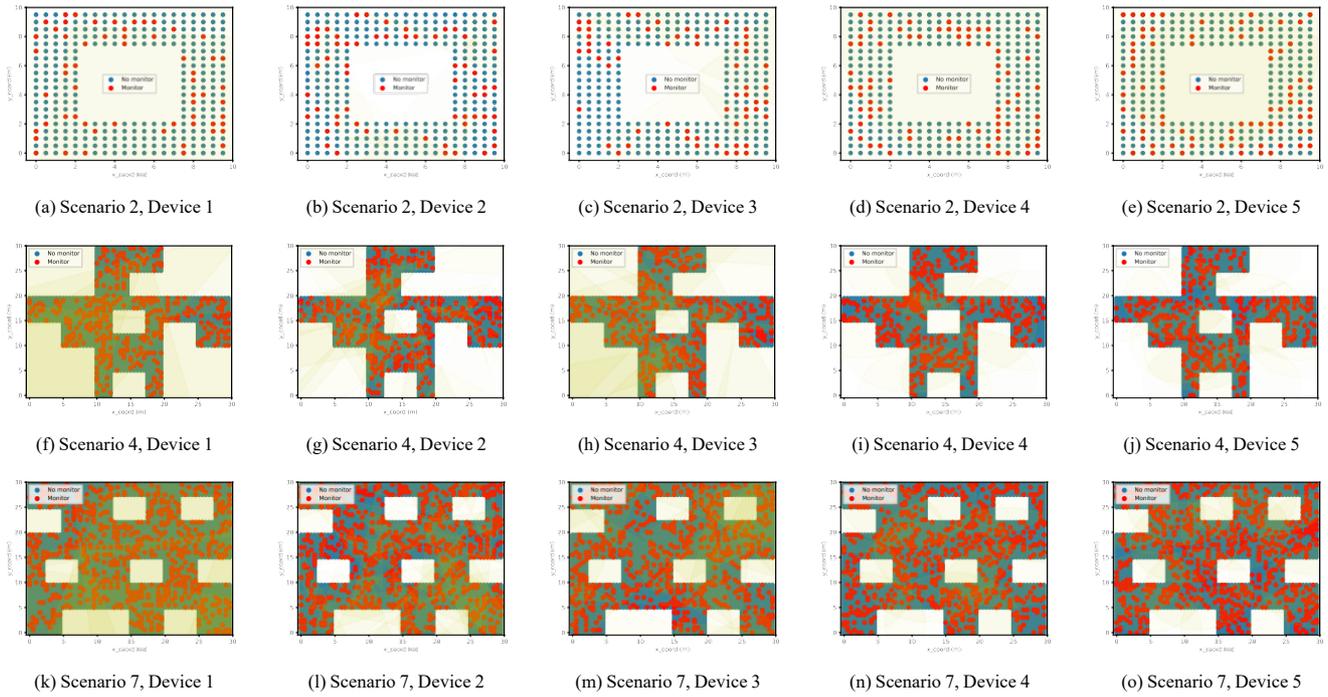

Figure 12: Best solutions obtained small size scenario (Scenario 2), medium size scenario (Scenario 4) and big size scenario (Scenario 7), detailed for all the devices considered. Yellow area indicates monitored area.